\documentclass[preprint, sort&compress]{elsarticle}
\journal{Computer Physics Communications}

\usepackage{amsmath}
\usepackage{amssymb}
\usepackage{bm}
\usepackage[nolabel]{showlabels}
\usepackage[ISO=false]{diffcoeff}
\usepackage{listings}
\usepackage{graphicx}
\usepackage[utf8]{inputenc}
\usepackage{tablefootnote}
\usepackage{threeparttable}
\usepackage{makecell}
\usepackage{xcolor}
\definecolor{mydarkgray}{gray}{0.2}

\usepackage[rightComments=false, endLComment=$$, commentColor=mydarkgray]{algpseudocodex}
\usepackage[plain]{algorithm}
\usepackage{tikz}
\usetikzlibrary{calc, arrows.meta}

\usepackage{geometry}

\newcommand{\bz}{\bm{z}}
\newcommand{\bphi}{\bm{\phi}}
\newcommand{\btheta}{\bm{\theta}}
\newcommand{\vphi}{\varphi}
\newcommand{\bvphi}{\bm{\vphi}}

\newcommand{\pacpt}{p_{a}}

\newcommand{\var}[1]{\operatorname{var}\left[#1\right]}

\newcommand{\E}[1]{E\left[#1\right]}

\newcommand{\g}{\mathbf{g}}
\newcommand{\qpt}{q(\bphi|\btheta)}
\diffdef{} 
{
    op-symbol = \mathrm{d},
    *-op-left = true,
    op-order-sep = 0 mu,
}

\diffdef{p}
{
    left-delim = \left .,
    right-delim = \right|,
    subscr-nudge = 1 mu
}
\begin{document}

\begin{frontmatter}

\title{Gradient estimators for normalising flows}
\author[iis]{Piotr Białas}
\author[ift]{Piotr Korcyl}
\author[ift]{Tomasz Stebel}

\address[iis]{Institute of Applied Computer Science, Jagiellonian University, ul.~Łojasiewicza 11, 30-348 Kraków Poland}
\address[ift]{Institute of Theoretical Physics, Jagiellonian University, ul.~Łojasiewicza 11, 30-348 Kraków Poland}

\begin{abstract}
 Recently a machine learning approach to Monte-Carlo simulations called Neural Markov Chain Monte-Carlo (NMCMC) is gaining traction. In its most popular form it uses neural networks to construct normalizing flows which are then trained to approximate the desired target distribution. In this contribution we present new gradient estimator for Stochastic Gradient Descent algorithm (and the corresponding \texttt{PyTorch} implementation) and show that it leads to better training results for $\phi^4$ model. For this model our estimator achieves the same precision in approximately half of the time needed in standard approach and ultimately provides better estimates of the free energy. We attribute this effect to the lower variance of the new estimator. In contrary to the standard learning algorithm our approach does not require estimation of the action gradient with respect to the fields, thus has potential of further speeding up the training for models with more complicated actions.
\end{abstract}

\end{frontmatter}

\section{Introduction}

Despite the apparent simplicity of the original idea behind Monte Carlo simulations by Stanislaw Ulam, this approach is one of the pillars of computational sciences. Expressed in a form of an algorithm applied to study a simple classical statistical mechanics problem by Metropolis et al. \cite{metropolis}, it is ubiquitous as a tool of dealing with complicated probability distributions (see for example \cite{binder}). In many cases one resorts to the construction of an associated Markov chain of consecutive proposals which provides a mathematically grounded way of generating samples from a given distribution even when the proper normalization of the latter is not known. The only limiting factor of the approach is the statistical uncertainty which directly depends on the number of statistically independent configurations. Hence, the effectiveness of any such simulation algorithm can be linked to its autocorrelation time which quantifies how many configurations are produced before a new, statistically independent configuration appears. For systems close to phase transitions the increasing autocorrelation times, a phenomenon called critical slowing down, is usually the main factor which limits the statistical precision of outputs. 

The recent interest in machine learning techniques has offered possible ways of dealing with this problem. 
Ref.~\cite{VANPRL} and later Ref.~\cite{PhysRevE.101.023304} proposed autoregressive neural networks as a mechanism of generating independent configurations which can be used as proposals in the construction of the Markov chain. The new algorithm was hence called Neural Markov Chain Monte Carlo (NMCMC). Once the neural network is sufficiently well trained one indeed finds that autocorrelation times are significantly reduced as was demonstrated in the context of the two-dimensional Ising model in Ref.~\cite{bialas2021analysis}.

For systems with continuous degrees of freedom the NMCMC algorithm has to be appropriately modified and the predominant approach is to use {\em normalizing flows} to generate configurations while at the same time calculating  their probabilities. Both of these steps are necessary. Before any neural network can be used for that purpose it must be trained, {\em i.e.} its weights should be tuned in such a way as to approximate the desired probability distribution. The standard approach for achieving this is using the stochastic gradient descent (SGD) algorithm which requires the estimation of gradients of the loss function with respect to the neural network weights. In this contribution we propose to adapt the gradient estimator used for  auto-regressive networks applied for discrete models (e.g. Ising model) to the case of normalizing flows. We show that this estimator avoids calculating the derivative of the action and is only approximately 10\% slower. Furthermore, because of its better convergence properties when applied in SGD, it  outperforms the standard algorithm in terms of resulting autocorrelation time  and the quality of calculations of variational free energy. We attribute this effect to the lower variance of this new estimator. We demonstrate our idea using a solvable toy-model and the scalar $\phi^4$ field theory by comparing the proposed gradient estimator with other two gradient estimators used in the Literature.

This contribution is organised as follows. In order to be self-contained we briefly introduce the NMCMC approach in Section~\ref{sec:nmcmc}. Then we describe different gradient estimators in Section~\ref{sec:grad-estimators}. We provide the definitions, as well some characteristics.  In Section~\ref{sec:elim} we discuss how the estimators introduced in Section~\ref{sec:grad-estimators} can be adapted to work with normalising flows. Section~\ref{sec:toy-model} provides a very simple toy example, where we can thoroughly compare all estimators. Finally, in Section~\ref{sec:phi4}, we compare all the estimators on the two dimensional $\phi^4$ model. We also include the snippets of Python code that implement our estimator using \texttt{PyTorch} framework\cite{PyTorch}. 

\section{Neural Markov Chain Monte Carlo}
\label{sec:nmcmc}

When using the  Monte-Carlo methods we are faced with the task of generating samples from some {\em target} distribution $p(\bphi)$. In the majority of the interesting applications, {\em e.g.}  lattice field theories, it is impossible to generate samples  independently from this distribution and instead we have to resort to Markov Chain Monte Carlo methods (MCMC). 

In this approach given an initial configuration $\bphi_i$, a new trial configuration $\bphi_{trial}$ is proposed from the distribution $q(\bphi_{trial}|\bphi_i)$. This trial configuration is then accepted with probability $\pacpt(\bphi_{trial}|\bphi_i)$ or the previous configuration is repeated in the Markov chain. Usually  the configuration $\bphi_{trial}$ differs from $\bphi_i$ only on a small subset of degrees of freedom like {\em e.g.} single lattice site. If the so called {\em detailed balance} condition
\begin{equation}\label{eq:detailed-balance}
    p(\bphi_i)q(\bphi_{trial}|\bphi_i)\pacpt(\bphi_{trial}|\bphi_i) = p(\bphi_{trial})q(\bphi_{i}|\bphi_{trial})\pacpt(\bphi_{i}|\bphi_{trial})
\end{equation}
is satisfied and provided that all available configurations can be reached, then asymptotically this procedure generates samples with distribution $p(\bphi)$. One way of satisfying condition \eqref{eq:detailed-balance} is by Metropolis-Hastings acceptance probability
\begin{equation}
    \pacpt(\bphi_{trial}|\bphi_i)=\min\left\{1,\frac{p(\bphi_{trial})}{q(\bphi_{trial}|\bphi_i)}\frac{q(\bphi_{i}|\bphi_{trial})}{p(\bphi_{i})}\right\}
\end{equation}
The biggest drawback of this algorithm is the fact that consecutive samples are highly correlated due to small incremental changes made at each step. 

The idea of Metropolized Independent Sampling (MIS) method \cite{Liu} is to generate samples {\em independently} from some distribution $q(\bphi)$ {\em i.e.}
\begin{equation}\label{eq-detailed-balance}
q(\bphi_{trial}|\bphi_i) = q(\bphi_{trial})
\end{equation}
and then proceed with the Metropolis-Hastings accept/reject step,
\begin{equation}\label{eq:ims}
    \pacpt(\bphi_{trial}|\bphi_i)=\min\left\{1,
    \frac{p(\bphi_{trial})}{q(\bphi_{trial})}
    \frac{q(\bphi_i)}{p(\bphi_i)}\right\}.
\end{equation}
This also introduces correlations but if the distribution $q(\bphi)$ is sufficiently close to $p(\bphi)$ and the acceptance rate is close to one, then those correlations can be substantially smaller then in the case of MCMC (see Ref.~\cite{bialas2021analysis} for discussion). 

Seemingly, in the MIS approach one has only replaced the problem of generating configurations from the distribution $p(\bphi)$ with another hard problem of finding the distribution $q(\bphi)$ that is close to the target distribution $p(\bphi)$ and allows for fast generation of independent configurations. However, following the proposal of Neural Markov Chain Monte Carlo one can use Machine Learning techniques, notably neural networks, to {\em learn} the distribution $q(\bphi)$ \cite{VANPRL,PhysRevE.101.023304}. The general idea is that $q(\bphi)$ is now parameterized by some (very large) set of parameters $\btheta$
\begin{equation*}
    q(\bphi)=q(\bphi|\btheta).
\end{equation*}
The training consists in the tuning of the parameters $\theta$ as to  minimize a loss function  that measures the difference between  $q(\bphi|\btheta)$ and target distribution $p(\bphi)$.  A natural choice for such a function is the  Kullback–Leibler divergence
\begin{equation}\label{eq-KL}
    D_{KL}(q|p) = \int\dl\bphi \, q(\bphi|\btheta) \left(\log q(\bphi|\btheta)-\log p(\bphi)\right)= E[\log q(\bphi|\btheta)-\log p(\bphi)]_{q(\bphi|\btheta)}.
\end{equation}
Please note that this function is not symmetric: $D_{KL}(q|p)\neq  D_{KL}(p|q)$.
This particular form \eqref{eq-KL} is chosen because we are sampling from the distribution $q(\bphi|\btheta)$.

Actually, in most cases we know the target distribution $p(\bphi)$ only up to a normalizing constant. Let us assume that we only know $P(\bphi)$,
\begin{equation}
    P(\bphi)= Z \cdot p(\bphi),\qquad Z= \int\dl\bphi P(\bphi),
\end{equation}
where the constant $Z$ is usually called the {\em partition function}. Inserting $P$ instead of $p$ into Kullback-Leibler divergence definition we obtain the {\em variational free energy}
\begin{equation}\label{eq:shifted-DKL}
\begin{split}
   F_{q} = \E{\log q(\bphi|\btheta)-\log p(\bphi) -\log Z}_{q(\bphi|\btheta)}= F + D_{KL}(q|p),
\end{split}
\end{equation}
where $F=-\log Z$ is the {\em free energy}. As $F$ does not depend on $\btheta$, minimizing $F_q$ is equivalent to minimizing $D_{KL}$. In the following we will use $P$ and $F_q$ instead of $p$ and $D_{KL}$. The possibility of calculating $F_q$ and thus estimating the free energy $F$ is one of the major strengths of this approach as this is very hard to do in the classical MCMC simulations \cite{ETOinLFT}.

It is a non-trivial question as to how to define the $\qpt$ model in practice. It has to: 1) define a properly normalized probability distribution and 2) allow for  sampling from this distribution. We will shortly describe two common approaches: normalizing flows and autoregressive networks which can be used for systems with continuous and discrete degrees of freedom respectively. 

\subsection{Continuous degrees of freedom -- Normalizing flows}
\label{sec:normalising-flows}

The normalizing flow can be thought of as a tuple of functions \cite{dinh2017density,PhysRevD.100.034515,9089305}
\begin{equation}\label{eq-nf}
   \mathbb{R}^{D}\ni \bz\longrightarrow (q_{pr}(\bz),\bm{\vphi}(\bz|\btheta))\in (\mathbb{R},\mathbb{R}^{D}).
\end{equation}
The function $q_{pr}(\bz)$ is the probability density defining a {\em prior} distribution of random variable $\bz$. The function $\bvphi(\bz|\btheta)$ must be a {\em bijection} so if the input $\bz$ is drawn from  $q_{pr}(\bz)$ then the output $\bphi$ is distributed according to 
\begin{equation}\label{eq-q-phi}
    q(\bphi|\btheta)= q_z(z|\btheta) \equiv  q_{pr}(z)J(\bz|\btheta)^{-1},\quad \bphi=\bvphi(\bz|\btheta),
\end{equation}
where
\begin{equation}\label{eq_jac_def}
    J(\bz|\btheta)=\det \left(\diffp{\bvphi(\bz|\btheta)}{\bz}\right)
\end{equation}
is the determinant of the Jacobian of $\bvphi(\bz|\btheta)$. For this approach to be of practical use the flows are constructed in such a way that the Jacobian determinant is relatively easy to compute. 
In terms of $q_{pr}(\bz)$, $q_z(\bz|\btheta)$ and $\bvphi(\bz)$ the variational free energy $F_q$  can be written as
\begin{equation}
    F_q = \int\dl\bz\, q_{pr}(\bz) \left(\log q_z(\bz|\btheta)-\log P(\bvphi(\bz|\btheta))\right)=\E{\log q_z(\bz|\btheta)-\log P(\bvphi(\bz|\btheta))}_{q_{pr}(\bz)}.
\end{equation}
When sampling from $q_{pr}(\bz)$ this can be approximated as
\begin{equation}\label{eq:Fq-normalizing-flow}
      F_q \approx\frac{1}{N} \sum_{i=1}^N \left(\log q_z(\bz_i|\btheta)-\log P(\bvphi(\bz_i|\btheta))\right),\quad \bz_i\sim q_{pr}(\bz),
\end{equation}
where the $\sim$ symbol denotes that each $\bz_i$ is  drawn from the distribution $q_{pr}(\bz)$.

\subsection{Discrete degrees of freedom -- Autoregressive networks}
\label{sec:autoregressive}

For systems with discrete degrees of freedom we cannot use normalizing flows. In such situation we can represent the distribution $q(\bphi|\btheta)$ via conditional probabilities,  
\begin{equation}\label{eq:product-rule}
    q(\bphi|\btheta) = q(\phi_1|\btheta)\prod_{i=2}^{\mathcal{N}} q(\phi_i|\phi_{i-1},\ldots,\phi_1,\btheta),
\end{equation}
where  $(\phi_1,\ldots,\phi_{\mathcal{N}})$ are the $\mathcal{N}$ components of the configuration $\bphi$ and have to represent {\em discrete}  degrees of freedom. 
In the case of a simple spin system, $\phi_i=\pm 1$, and the factorised probability \eqref{eq:product-rule} can be described by a neural network with $\mathcal{N}$ inputs $\phi_1,\ldots,\phi_{\mathcal{N}}$ and $\mathcal{N}$ outputs corresponding to conditional probabilities $q(\phi_i=1|\phi_{i-1},\ldots,\phi_1)$.  This can be generalized to the case when $\phi_i$ takes on more then two values. 
To ensure that  $q(\phi_i|\phi_{i-1},\ldots,\phi_1,\btheta)$ depends  only on the values of the preceding spins $\phi_{i-1},\ldots,\phi_1$ the so called {\em autoregressive networks} are used  \cite{VANPRL, graphical_models, NeuralAutoregressive, MaskedAutoencoders, pixelCNN}.

The configuration $\bphi=(\phi_1,\ldots,\phi_{\mathcal{N}})$ can be generated by successively generating the components $\phi_i$  one by one from distributions $q(\phi_i|\phi_{i-1},\ldots,\phi_1,\btheta)$ starting at $\phi_1$ and feeding them successively back to the network to obtain $q(\phi_{i+1}|\phi_{i},\ldots,\phi_1,\btheta)$. 

In this formulation  we can obtain an estimate of $F_q$ by sampling $\bphi$ directly from $q(\bphi|\btheta)$,
\begin{equation}\label{eq:Fq-autoregressive}
    \begin{split}
    F_q\approx\frac{1}{N} \sum_{i=1}^N \left(\log q(\bphi_i|\btheta)-\log P(\bphi_i)\right),\quad \bphi_i\sim q(\bphi|\btheta).
\end{split}
\end{equation}

\section{Gradient estimators}
\label{sec:grad-estimators}

Minimizing $F_q$ and thus training the machine learning model is done be the stochastic gradient descent (SGD) and requires the calculation of the gradient of $F_q$ with respect to $\btheta$. Actually, we can only estimate the gradient based on the finite sample (batch) of $N$ configurations $\{\bphi\} = \{\bphi_1,\ldots,\bphi_N\}$. 

In the case of normalizing flows this is pretty straightforward. We can directly differentiate expression \eqref{eq:Fq-normalizing-flow} to obtain the first gradient estimator $\g_3[\{\bphi\}]$,
\begin{equation}\label{eq:g1}
   \begin{split}
        \diff{F_q}{\btheta} &\approx \g_3[\{\bphi\}]
        \equiv\frac{1}{N}\sum_{i=1}^N\diff*{\left(\log q(\bz_i|\btheta)-\log P(\bvphi(\bz_i|\btheta))\right)}{\btheta},\quad\bz_i\sim q_{pr}(\cdot|\btheta).
   \end{split}
\end{equation}
This derivative can  be  calculated by popular packages like {\em e.g.} \texttt{PyTorch}\cite{PyTorch} or \texttt{TensorFlow}\cite{tensorflow2015-whitepaper} using automatic differentiation.

While conceptually simple, this estimator has a considerable drawback as it requires calculating the gradient of the distribution $P(\bphi)$ with respect to the configuration $\bphi$,
\begin{equation*}
  \diffp*{\log P(\bvphi(\bz_i|\btheta))}{\btheta} = \diffp*{\log P(\bphi)}{\bphi}[\bphi=\bvphi(\bz_i|\btheta)] \diffp{\bvphi(\bz_i|\btheta)}{\btheta}. 
\end{equation*}
In lattice field theories the probability $P$ is given by the {\em action} $S(\bphi)$,
\begin{equation}
    \log  P(\bphi(\bz|\btheta))=-S(\bphi(\bz|\btheta)) 
\end{equation}
and so calculating the gradient of $F_q$ requires the gradient of the action $S$ with  respect to the  fields $\bphi$. This may not pose large complications for {\em e.g.} $\phi^4$ theory where action is just a polynomial in $\bphi$. Other lattice field theories however, notably the Quantum Chromodynamics with dynamical fermions, may have much more complicated actions including some representation of the nonlocal determinant of the fermionic matrix and the calculation of the action gradient may be impractical.

Autoregressive networks require calculating the derivative of expression \eqref{eq:Fq-autoregressive}. This is more tricky as in this case the sampling distribution $\qpt$ also depends on $\btheta$. Following Ref.~\cite{VANPRL} we can however start by calculating the gradient of the exact expression \eqref{eq:shifted-DKL},
\begin{equation}
\begin{split}
\label{eq:Fq-grad-autoregessive}
 \diff{F_q}{\btheta} & = \int\dl\bphi \, \diffp{q(\bphi|\btheta)}{\btheta} \left(\log q(\bphi|\theta)-\log P(\bphi)\right) \\
  &\phantom{=}+\int\dl\bphi \,  q(\bphi|\btheta) \diffp*{\log q(\bphi|\theta)}{\btheta}.
  \end{split}
\end{equation}
The last term in the above expression is zero because it can be rewritten as the derivative of a constant,
\begin{equation}\label{eq-const-der}
  \E{\diffp{\log q(\bphi|\theta)}{\btheta}}_{\qpt} = \int\dl\bphi \,  \diffp{ q(\bphi|\theta)}{\btheta} =\diffp*{ \underbrace{\int\dl\bphi \,   q(\bphi|\theta)}_1}{\btheta}  = 0.
\end{equation}
First term in expression \eqref{eq:Fq-grad-autoregessive} can be further rewritten as
\begin{equation}\label{eq-KL-grad-simpl}
\begin{split}
     \diff{F_q}{\btheta} 
        & =\int\dl\bphi \, q(\bphi|\btheta)\diffp{\log q(\bphi|\btheta)}{\btheta} \left(\log q(\bphi|\theta)-\log P(\bphi)\right)\\
        &=\E{\diffp{\log q(\bphi|\btheta)}{\btheta}\left(\log q(\bphi|\theta)-\log P(\bphi)\right)}_{q(\bphi|\btheta)}
     \end{split},
\end{equation}
and approximated as
\begin{equation}\label{eq-KL-grad-approx}
   \diff{F_q}{\btheta}\approx \g_1[\{\bphi\}]\equiv\frac{1}{N} \sum_{i=1}^N \diffp{ \log q(\bphi_i|\btheta)}{\btheta} \left(\log q(\bphi_i|\btheta)-\log P(\bphi_i)\right),
\end{equation}
which defines another gradient estimator $\g_1[\{\bphi\}]$ discussed in this work.

The Authors of Ref.~\cite{VANPRL}  introduce yet another gradient estimator, which we label by $\g_2[\{\bphi\}]$, with the aim of reducing the variance, by subtracting the batch mean from the signal
\begin{equation}\label{eq:def-g2}
   \g_2[\{\bphi\}]= 
   \frac{1}{N}\sum_{i=1}^N 
   \diffp{ \log q(\bphi_i|\btheta)}{\btheta}  \left(s(\bphi_i|\btheta)-\overline{s(\bphi|\btheta)_N} \right),
\end{equation}
where 
\begin{equation}\label{eq:signal}
     s(\bphi|\btheta) \equiv  \log q(\bphi|\btheta)-\log P(\bphi)\quad\text{and}\quad  \overline{s(\bphi|\btheta)_N} =\frac{1}{N}\sum_{i=1}^{N} s(\bphi_i|\btheta).
\end{equation}
Please note that expressions (\ref{eq-KL-grad-approx}) and (\ref{eq:def-g2}) do not depend on $Z$.

Contrary to $\g_3$ and $\g_1$, the $\g_2$ estimator is slightly  biased
\begin{equation}\label{eq:g2-bias}
\E{\g_2[\{\bphi\}]}=\frac{N-1}{N}\E{\g_1[\{\bphi\}]}.
\end{equation}
The proof of this fact is presented in \ref{ap:g2}. Of course such multiplicative bias does not play any role when the estimator is used in the gradient descent algorithm and is very small anyway when $N\sim 10^3$. For all practical purposes we can treat all estimators as unbiased, 
so any differences must stem from the higher moments, most importantly from the variance. 

Although not much can be said about the variances of these estimators in general, we can show that for perfectly trained model {\em i.e.} when $\qpt=p(\bphi)$,
\begin{equation}\label{eq:g1-variance}
   \var{\g_1[\{\bphi\}}_{\qpt=p(\bphi)}=\frac{1}{N}(\log Z)^2 \var{ \diffp{ \log q(\bphi|\btheta)}{\btheta} }_{\qpt=p(\bphi)}
\end{equation}
As $Z$ may be very large or very small depending on formulation of $P(\bphi)$, the variance can be quite substantial. 
For $\g_2$ we obtain
\begin{equation}
   \var{g_2[\{\bphi\}}_{\qpt=p(\bphi)}=0
\end{equation}
The proof is presented in  \ref{ap:variance}. This potentially very large reduction in variance was the actual  rationale for introducing this estimator (see Reference~\cite{VANPRL} supl. materials). 

As for estimator $\g_3$ we cannot make any claims as to the value of its variance even for  $\qpt=p(\bphi)$ but we will show that it does not need to vanish in this case. 

\section{Eliminating action derivative}
\label{sec:elim}

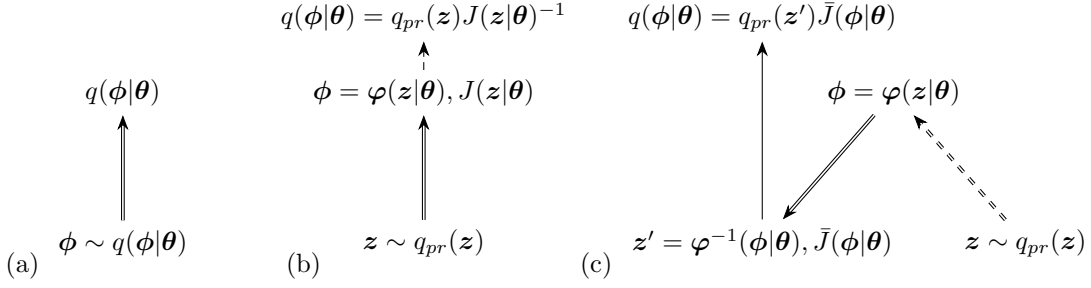
\begin{figure}
\begin{center}
\begin{tikzpicture}[>={Stealth[length=2mm]}]
\newlength{\yoffset}
\setlength{\yoffset}{0cm}
\newlength{\lbloffset}
\setlength{\lbloffset}{3mm}
\tikzstyle{flow}=[double]
\begin{scope}[yshift=\yoffset]
\node (phi) at (0, 0)  [rectangle, minimum size=5mm] {$\bphi\sim\qpt$};
\node (qpt) at (0, 2)  [rectangle, minimum size=5mm] {$\qpt$}; 
\draw[->, flow]    (phi) -- (qpt) {};
\draw  let \p1=(phi.west) in (\x1-10pt,-\yoffset-\lbloffset) node {(a)};
\end{scope}

\begin{scope}[xshift = 4cm, yshift=\yoffset]
\node (z) at (0,0)  [rectangle,minimum size=5mm, align=center] {$\bz\sim q_{pr}(\bz)$};
\node (qz) at (0, 2)  [rectangle, minimum size=5mm] {$\bphi=\bvphi(\bz|\btheta), J(\bz|\btheta)$}; 
\node (qpt) [above of = qz, node distance=1cm] {$\qpt=q_{pr}(\bz)J(\bz|\btheta)^{-1}$};
\draw[->, densely dashed] (qz)--(qpt);
\draw  let \p1=(qz.west) in (\x1,-\yoffset-\lbloffset) node {(b)};
\draw[->, flow]   (z) -- (qz);
\end{scope}     

\begin{scope}[xshift=9.5cm, yshift=\yoffset]
\node (zp) at (-1cm,0)  [rectangle, minimum size=5mm] {$\bz'=\bvphi^{-1}(\bphi|\btheta), \bar J(\bphi|\btheta)$};
\node (phi) at (0.75cm,2)  [rectangle, minimum size=5mm] {$\bphi=\bvphi(\bz|\btheta)$};
\node (z) at (2.5,0) [rectangle, minimum size = 5mm] {$\bz\sim q_{pr}(\bz)$};
\node (qpt) [above of = zp, node distance=30mm] {$\qpt=q_{pr}(\bz')\bar J(\bphi|\btheta)$};
\draw[->] (zp)--(qpt);
\draw[->, flow]   (phi) -- (zp) ;
\draw[->, dashed, flow]   (z) -- (phi);
\draw  let \p1=(qpt.west) in (\x1-3mm,-\yoffset-\lbloffset) node {(c)};
\end{scope}    
\end{tikzpicture}
\end{center}
\caption{Schematic picture of three algorithms for gradient estimation discussed in the paper: a) autoregressive networks, b) normalizing flows c) our proposition of adaptation of a) into normalizing flows. Double line arrows represent the flow: upward-pointing arrows represents forward propagation, downward-pointing arrows represents backward propagation. Dashed arrows denote propagation which doesn't require gradient.}
\label{fig_schem_alg}
\end{figure}

We notice that contrary to $\g_3$, the estimators $\g_1$ and $\g_2$ do not require calculating the derivatives of $P(\bphi)$. This is due to the fact that we can first generate a configuration $\bphi$ from the distribution $q(\bphi|\btheta)$ and then obtain its probability directly (see Figure~\ref{fig_schem_alg}a for schematic picture). In case of normalizing flows in the standard approach (estimator $\g_3$ described above) we do not have direct access to the function $q(\bphi|\btheta)$ since the probability of the configuration is determined simultaneously with generation, by passing $\bz$ through the network (see  Figure~\ref{fig_schem_alg}b). 
However, by leveraging the reversibility of normalizing flows we can adapt the $\g_2$ estimator to that case (see Figure~\ref{fig_schem_alg}c).  
The $\g_2$ estimator requires the $\qpt$ function, and while it is not explicit in the normalizing flows formulation \eqref{eq-nf}, it can be inferred from 
Eq.~\eqref{eq-q-phi}. 
Using the fact that the Jacobian determinant of transformation $\bvphi^{-1}(\bphi|\btheta)$ 
\begin{equation*}
 \bar J(\bphi|\btheta)\equiv \det \left(\diffp{\bvphi^{-1}(\bphi|\btheta)}{\bphi}\right)
\end{equation*}
is the inverse of Jacobian determinant of $\bvphi(\bz|\btheta)$, 
\[
\bar J(\bphi|\btheta)= J(\bz|\btheta)^{-1},
\]
we can write $\qpt$ as
\begin{equation}\label{eq:q-phi-inv}
\qpt=q_{pr}(\bz')\bar J(\bphi|\btheta)\quad \bz'=\bvphi^{-1}(\bphi|\btheta).
\end{equation}
Given that, the calculation of $\g_2$ would proceed as follows:
\begin{enumerate}
    \item  First use  the function $\bvphi(z|\theta)$ to generate configurations $\bphi_i$  without any gradient calculations.
    \begin{equation}
\bphi_i = \bvphi(\bz_i | \btheta)\quad \bz_i \sim q_{pr}(\bz)
\end{equation}
\item Then  switch on the gradient calculations and calculate  $\bz'$ by running the flow backward
\begin{equation*}
    \bz'_i=\bvphi^{-1}(\bphi_i|\btheta)
\end{equation*}
then use the Eq.~\eqref{eq:q-phi-inv} to calculate the probability $\qpt$.
It is very important that we use the $\bz'_i$ from step two and not $\bz_i$ from step one, as the gradients have to propagated through $q_{pr}$. 
\item And finally the gradient estimate is calculated as in \eqref{eq:def-g2} 
    \begin{equation*}
    \g_2[\{\bphi\}]=\frac{1}{N} \sum_{i=1}^N \diffp{ \log q(\bphi_i|\btheta)}{\btheta} 
   \left(\log q(\bphi_i|\btheta)-\log P(\bphi_i) -\overline{\log q(\bphi|\btheta)-\log P(\bphi)}\right).
\end{equation*}
\end{enumerate}
This will require running the flow two times: forward to obtain $\bphi_i$, then backward to calculate $\bz'$, but the gradients have to be calculated only on the  last pass. We illustrate this with  pseudocode in Algorithm~\ref{alg:g2} and  schematically in Figure~\ref{fig_schem_alg}c. 

\begin{algorithm}
\begin{center}
\caption{\label{alg:g2}Calculation of $\g_2$ estimator for normalizing flows. The resulting $loss$ can be used for automatic differentiation.} 
\begin{minipage}{\textwidth-1cm}
\begin{algorithmic}[0]
\LComment{generate $\bphi$}
\State Switch off gradient calculations
\State $\bz\sim q_{pr}(\bz)$
\State $\bphi \gets \bvphi(\bz|\btheta)$
\Comment{Forward pass}
\LComment{Calculate signal}
\State $s \gets \log q(\bphi|\btheta)-\log P(\bphi)$
\LComment{Calculate $\g_2$}
\State Switch on gradient calculations
\State $\bz'\gets \bvphi^{-1}(\bphi|\btheta)$
\Comment{Backward pass}
\State $q\gets q_{pr}(\bz'|\btheta)\det \left(\diffp{\bvphi^{-1}(\bphi|\btheta)}{\bphi}\right)$ 
\State $loss \gets \log q \times (s-\bar{s})$
\Comment{Forward pass}
\end{algorithmic}
\end{minipage}
\end{center}
\end{algorithm}

\section{Toy model}
\label{sec:toy-model}
We will illustrate the concepts introduced in previous sections with a very simple, one dimensional normalizing flow that generates an exponential distribution,
\begin{equation}\label{eq-nf-exp}
   (q_{pr}(z),\vphi(z|\theta))= \left(1,-\frac{1}{\theta}\log (1-z)\right ),\quad z\in [0,1).
\end{equation}
This will allow us to explicitly calculate the form of each estimator $\g_i$, as well as its variance.  
Using \eqref{eq-q-phi} we obtain
\[
q_z(z|\theta) = 1\cdot J(\bz|\btheta)= \theta(1-z).
\]
Combining this with the inverse flow, 
\[
z=\varphi^{-1}(\phi|\theta) \equiv 1-e^{-\phi\theta},
\]
we get  the  exponential distribution,
\begin{equation*}
    q(\phi|\theta) \equiv q_z(\varphi^{-1}(\phi|\theta)|\theta)=\theta e^{-\theta \phi}. 
\end{equation*}
The Jacobian determinant for the inverse flow is
\begin{equation*}
    \bar J(\phi|\theta)=\theta e^{-\phi\theta},
\end{equation*}
so using \eqref{eq:q-phi-inv} we get same result for $\qpt$.  

Given the target distribution,
\begin{equation*}
    p(\phi) = \lambda e^{-\lambda \phi}\quad\text{and}\quad P(\phi)=Z\cdot p(\phi), 
\end{equation*}
the  free energy can be easily calculated as
\begin{equation}
\begin{split}
    F_q  &=\theta \int\!\dl\phi\, e^{-\theta \phi}
    \left(\log \theta  -\log \lambda -\log Z-\phi\left(\theta -\lambda\right)\right)\\
    &= \log\theta - \log\lambda -\log Z-\frac{1}{\theta}\left(\theta-\lambda\right),
\end{split}    
\end{equation}
as well as its gradient,
\begin{equation}\label{eq-nf-exp-KL-grad-exact}
    \diff{F_q}{\theta} = \frac{1}{\theta^2}(\theta-\lambda).
\end{equation}

We calculate the gradient estimator $\g_1$  for this distributions and obtain
\begin{equation}
\begin{split}
   \g_1[\{\phi\}] & = \frac{1}{N} \sum_{i=1}^N \left(\frac{1}{\theta} -\phi_i\right)\left(\log\theta - \log \lambda-\log Z -\phi_i(\theta  -\lambda)\right)\\
   & = \frac{1}{\theta}(\log\theta-\log\lambda-\log Z) + (\theta-\lambda)\frac{1}{N}\sum_{i=1}^N\phi_i^2\\
   &\phantom{=}- \left[\frac{1}{\theta}(\theta-\lambda)+(\log\theta-\log\lambda-\log Z)\right]\frac{1}{N}\sum_{i=1}^N\phi_i .
\end{split}
\end{equation}
Since 
\begin{equation*}
    \E{\frac{1}{N}\sum_i\phi_i}=\frac{1}{\theta}\quad\text{and}\quad \E{\frac{1}{N}\sum_i\phi_i^2}=\frac{2}{\theta^2},
\end{equation*}
we obtain the correct expression \eqref{eq-nf-exp-KL-grad-exact} for $\E{\g_1}$ which means that the estimator is unbiased as expected.  The calculation of the variance is more involved and the final result is
\begin{equation}
    \begin{split}
        \var{\g_1} 
        &=\frac{13}{N}\frac{(\theta-\lambda)^2}{\theta^4}-\frac{6}{N}\frac{(\theta-\lambda)(\log \theta-\log \lambda-\log Z)}{\theta^3}
        +\frac{1}{N}\frac{(\log \theta-\log \lambda-\log Z)^2}{\theta^2}.
    \end{split}
\end{equation}
So, for $\theta=\lambda$, 
\begin{equation*}
    \var{\g_1}_{\theta=\lambda}=\frac{1}{N}\frac{(\log Z)^2}{\lambda^2},
\end{equation*}
which is non zero in case $Z\neq 1$ and can be arbitrarily large. 

For the  estimator $\g_2$  we have
\begin{equation}
\begin{split}\label{eq-grad-exp-red}
   \g_2[\{\phi\}]
   &= (\theta  -\lambda)\frac{1}{N} \sum_i \left( \phi_i - \frac{1}{\theta}\right) (\phi_i-\bar\phi_N),\quad  \bar\phi_N =\frac{1}{N} \sum_{j=1}^N \phi_j.
\end{split}   
\end{equation}
Using the relations
\begin{equation}
    \E{\bar\phi_N}=\E{\phi}=\frac{1}{\theta}
\end{equation}
and
\begin{equation}
    \frac{1}{N}\sum_{i=1}^N\E{\left(\phi-\frac{1}{\theta}\right)\bar\phi_N}
    =\frac{1}{N}\E{\left(\phi-\frac{1}{\theta}\right)\phi}+\frac{N-1}{N}\E{\left(\phi-\frac{1}{\theta}\right)}\E{\phi}
\end{equation}
we obtain that the expectation value of  estimator $\g_2$ is
\begin{equation}
    \E{ \g_2}=\frac{N-1}{N}(\theta-\lambda)\frac{1}{\theta^2},
\end{equation}
as predicted by Eq.~\eqref{eq:g2-bias}. The calculations of the variance are tedious and we present them to first order in $N^{-1}$,
\begin{equation}
    \var{\g_2} = \frac{7}{N}\frac{(\theta-\lambda)^2}{\theta^4}+(\theta-\lambda)^2 
    O\left(\frac{1}{N^2}\right).
\end{equation}

And finally  for estimator $\g_3$ we obtain
\begin{equation}
\begin{split}
     \g_3[\{\phi\}] &= \frac{1}{N}\sum_i \diff*{\left(\log\theta -\log \lambda +\log (1-z_i)\left(1  -\frac{\lambda}{\theta}\right)\right)}{\theta}\\
     & = \frac{1}{\theta}+\frac{\lambda}{\theta^2}\frac{1}{N}\sum_i \log (1-z_i).
\end{split}
\end{equation}
Because  the random  variable $-\log(1-z)$ is distributed according to the exponential distribution with mean equal to one,
\begin{equation}
    E[\log(1-z_i)]=-1\quad\text{and}\quad \var{\log(1-z_i)}=1,
\end{equation}
we again obtain the correct result \eqref{eq-nf-exp-KL-grad-exact} for $\E{\g_3}$. Similarly variance can be calculated as  
\begin{equation}
    \var{\g_3} = \frac{1}{N}\frac{\lambda^2}{\theta^4}. 
\end{equation}
Please note that this expression  does not vanish when $\theta=\lambda$.

\subsection{Numerical results}

\begin{figure}
    \begin{center}
    \includegraphics[width=0.49\linewidth]{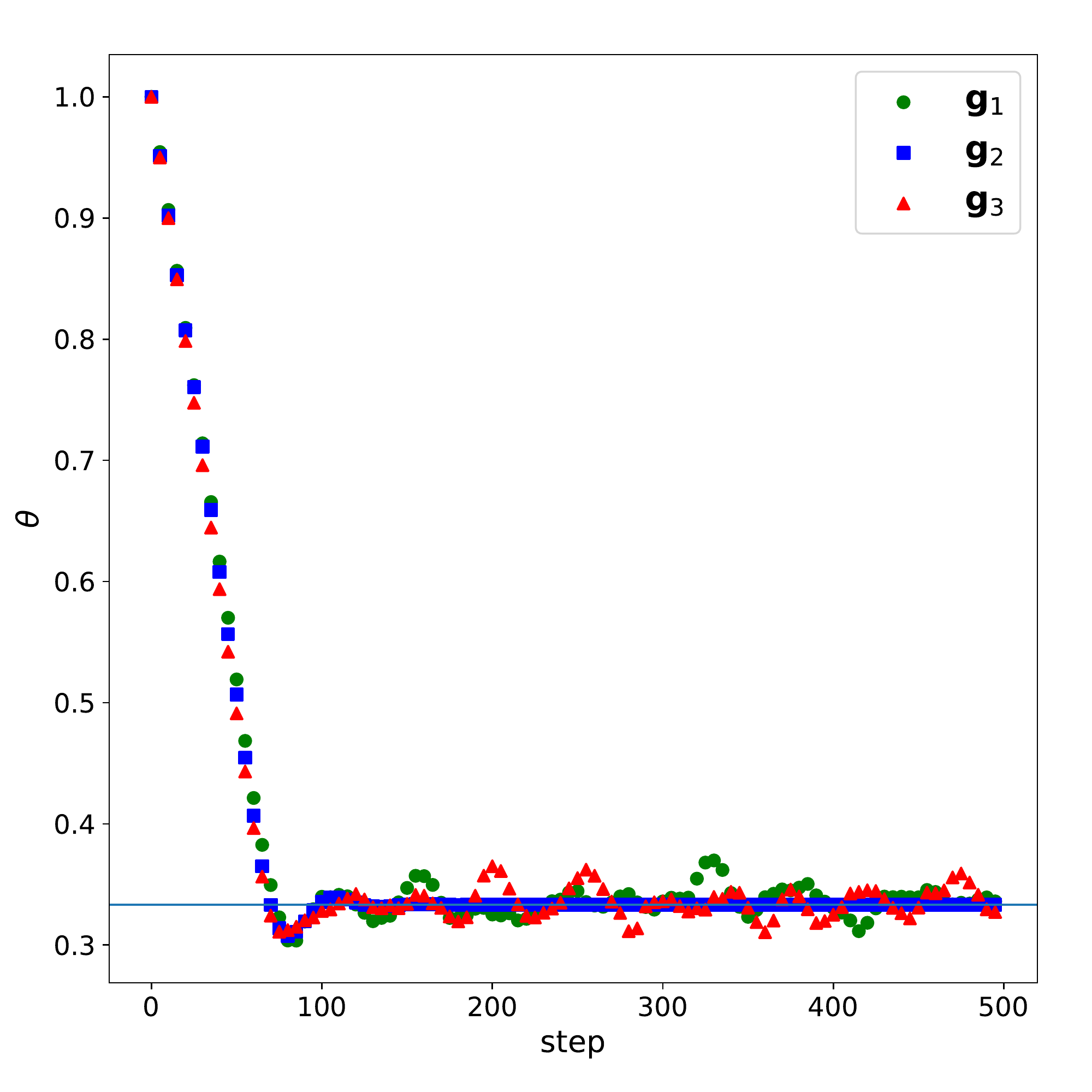}\includegraphics[width=0.49\linewidth]{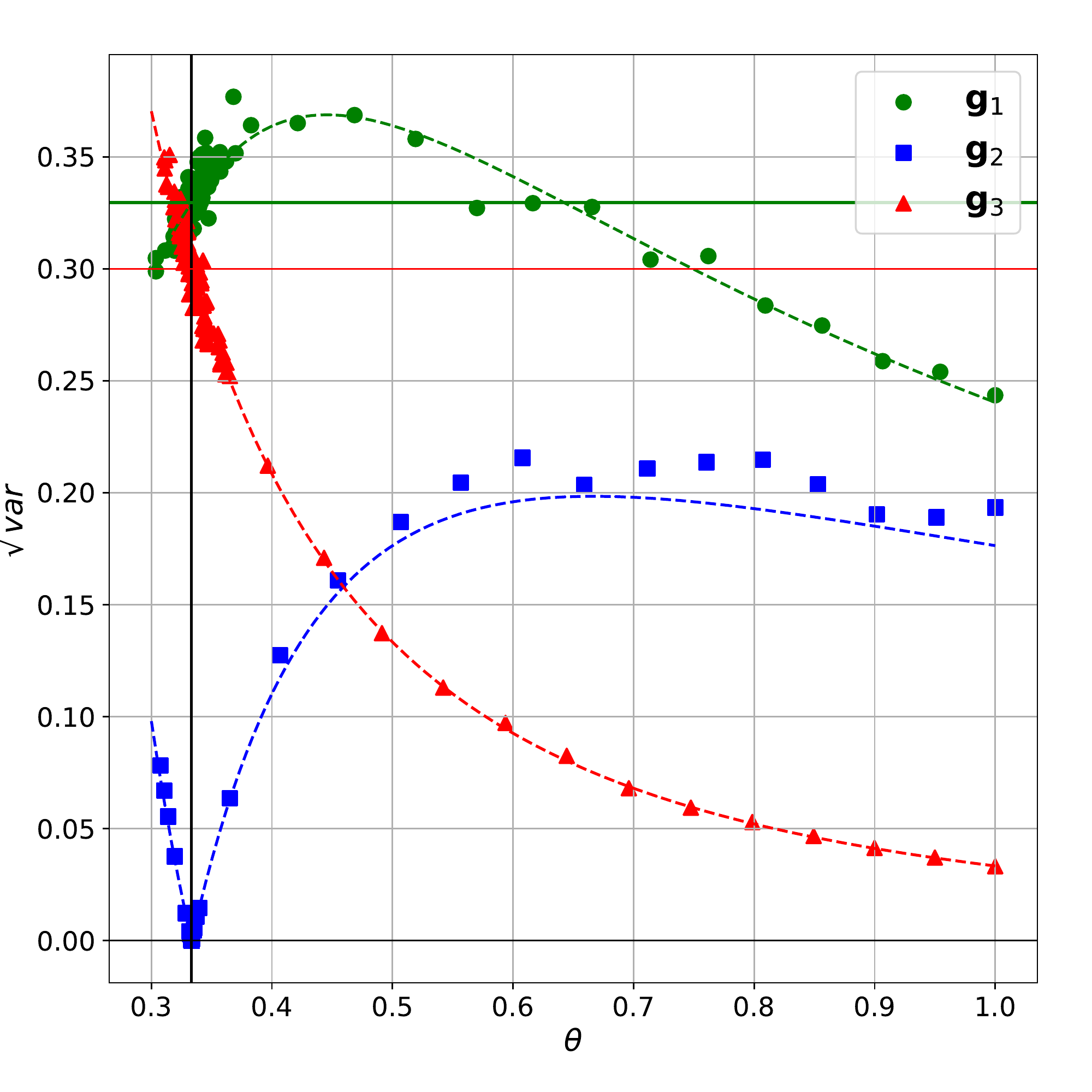}
    \end{center}
    \caption{\label{fig-toy}Optimizing the $q(\cdot|\theta)$ distribution. (Left) Evolution of parameter $\theta$, the blue horizontal line indicates the true value $\theta=1/3$. (Right) Standard deviation (square root of variance) of the gradient estimator vs. $\theta$, dotted lines represent analytic results. Vertical black line correspond to $\theta=1/3$ and horizontal lines correspond to values of standard deviation at $\theta=\lambda$ for $\g_1$ and $\g_3$ estimators.}
\end{figure}

In order to see how these three different estimators behave when employed in the SGD algorithm we have optimised the model $q(\bphi|\theta)$ to match the distribution $p(\btheta)$ using the \texttt{PyTorch} framework \cite{PyTorch}. The target distribution parameter $\lambda$ was set to $1/3$ and $Z$ to $\lambda^{-1}$. The starting $\theta$ value was set to 1. We have performed 500 steps, where by one step we understand a single update of the parameter $\theta$. At each step we have sampled a batch of $N=100$ elements from the distribution  $q(\phi|\theta)$ which we have used to calculate the gradient estimate  using one of the $\g_i$ estimators. The actual step {\em i.e.} adjustment of $\theta$ was performed using the Adam optimizer with learning rate set to 0.01. The results are presented in the left panel of Figure~\ref{fig-toy}. As we can see  all estimators  give similar performance and $\theta$ converges to the true value. However, after approximate convergence we note that the use of the estimators $\g_1$ and $\g_3$ results in rather large ``wandering'' of the $\theta$ around its target value,  which is due to  non-vanishing variance of the gradient in this case (see the right panel).

To estimate the variance we have generated 1000 additional batches at each step. On each batch we have calculated the gradient estimator and used those 1000 samples to estimate the variance. The results are presented in the right panel of Figure~\ref{fig-toy} where we show the standard deviation (square root of variance) of gradient estimators calculated during simulations. They are consistent with our analytical calculations and, as predicted, the variance of estimator $\g_2$ does vanish as $\theta\rightarrow \lambda$. In contrast the variance of the remaining estimators is substantially bigger then zero. While this is a contrived example it  serves as an indicator that while unbiased, different estimators can have dramatically different statistical properties. Of course increasing the batch size would result in decreased variance for all estimators.

\section{Lattice $\phi^4$ theory}
\label{sec:phi4}

The second example is the two dimensional scalar $\phi^4$ field theory with Euclidean action
\begin{equation}
S[\bphi|m^2,\lambda] =    \int\dl x^2 \left(
    \sum_{\mu=0,1}(\partial_\mu\phi(x))^2+m^2\phi^2(x)+\lambda\phi^4(x)
    \right)
\end{equation}
which, following \cite{albergo2021introduction}, we discretize as
\begin{equation}
\begin{split}
    S(\bphi|m^2,\lambda) &= \sum_{i,j=0}^{L-1}
    \phi_{i,j}\left( 
    2\phi_{i,j}-\phi_{i-1,j} -\phi_{i+1,j} + 2\phi_{i,j}-\phi_{i,j-1} -\phi_{i,j+1}\right)\\
    &\phantom{=}+\sum_{i,j=0}^{L-1}\left(m^2\phi_{i,j}+\lambda\phi_{i,j}^4\right),
\end{split}    
\end{equation}
where the lattice has size $L\times L$.
The probability distribution $p$ is given by the Boltzmann distribution
\begin{eqnarray}\label{eq:phi4-prob}
p(\bphi)=Z(m^2,\lambda)^{-1}e^{-S(\bphi|m^2,\lambda)}\qquad Z(m^2,\lambda)=\int\dl\bphi e^{-S(\bphi|m^2,\lambda)}
\end{eqnarray}
so $P(\bphi)=\exp(-S(\bphi))$.

\begin{figure}
    \begin{center}
    \includegraphics[width=0.5\linewidth]{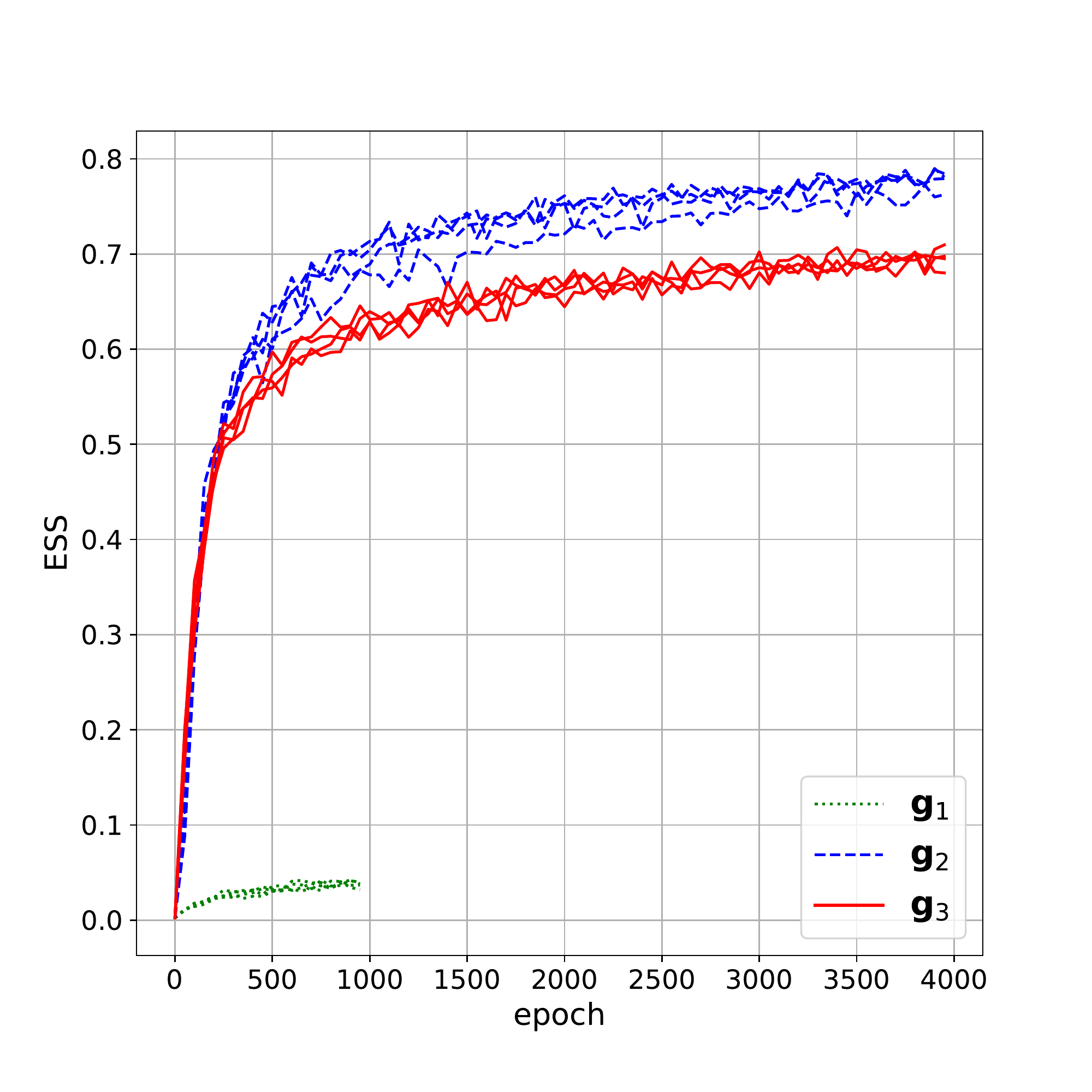}\includegraphics[width=0.5\linewidth]{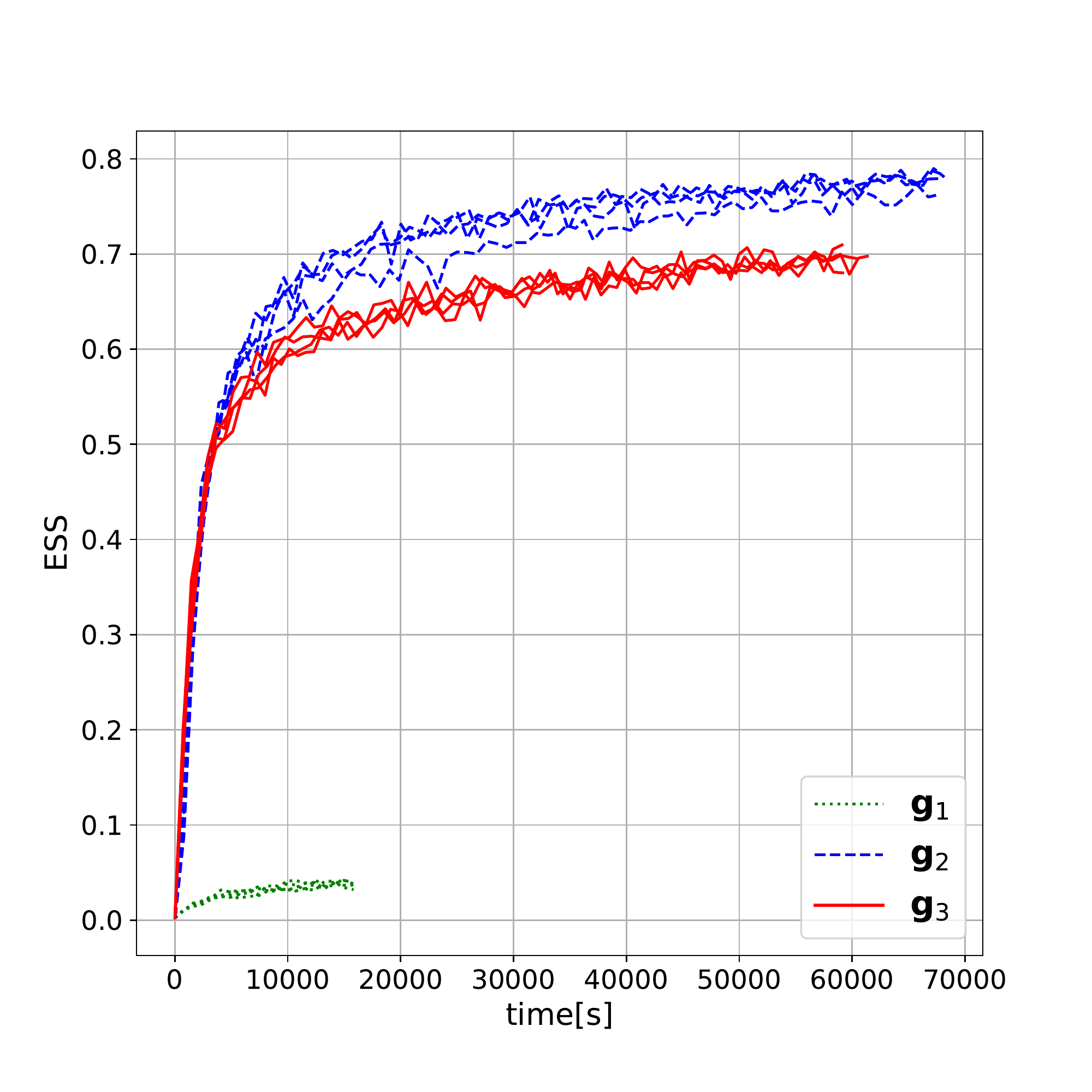}
    \end{center}
    \caption{\label{fig:phi4-ess}(Left) ESS as the function of epoch. One epoch consisted of 100 steps. In each step the \lstinline{train_step} function was called once.  (Right) ESS as a function of wall time in seconds. }
\end{figure}

We have used  \lstinline{PyTorch} normalizing flows implementation provided in the excellent tutorial \cite{albergo2021introduction}. It uses the {\em affine coupling layers} to implement the flow \cite{dinh2017density}. Field $\bphi$ is split using a checkerboard pattern into two parts $\bphi_1$ and $\bphi_2$. Part $\bphi_2$ is frozen and does not change during the transformation but is used as an input to functions $\bm t$ and $\bm s$ which are then used to transform part $\bphi_1$
\begin{equation}
\begin{split}
\bphi'_1 &\leftarrow \bphi_1 e^{\displaystyle \bm s(\bphi_2)}+\bm t(\bphi_2)   \\ 
\bphi'_2 &\leftarrow \bphi_2.
\end{split}
\end{equation}
The outputs of functions $\bm t$ and $\bm s$ have same size  as $\bphi_1$ and all arithmetic operations are performed pointwise. 
The Jacobian determinant of this transformation is easily calculable,
\[
\log J(\bz|\btheta)=\sum_{i}s_i(\bphi_2),
\]
where the sum runs over all components of $\bm s$. 

Please note that this is a bijection with the inverse transformation given by
\begin{equation}
\begin{split}
\bphi_1 &\leftarrow (\phi_1'-\bm t(\phi'_2))e^{\displaystyle -\bm s(\bphi'_2)}   \\ 
\bphi_2 &\leftarrow \bphi'_2.
\end{split}
\end{equation}
and 
\[
\log \bar J(\bphi|\btheta)=-\sum_{i}s_i(\bphi'_2)
\]

In the next layer parts  $\bphi_1$ and $\bphi_2$ are interchanged. Functions $\bm t$ and $\bm s$ in each layer are implemented using 
a convolutional neural network with two output channels. The architecture of this network is presented in table~\ref{tab-cnn}.  We use 16 coupling layers.
The prior distribution $q_{pr}$ is  taken as the standard normal distribution  $\mathcal{N}(0,1)$ independently on each component of $\bz$.

\begin{table}
    \centering
    \begin{tabular}{|c|rrc|}
\hline
layer    & $\text{ch}_{in}$ & $\text{ch}_{out}$ & kernel \\\hline\hline
     1 &  1   & 16 & (3,3) \\\Xcline{2-4}{0.1pt}
       &  \multicolumn{3}{c|}{\small Leaky ReLU}\\\hline
     2 &  16  & 16 & (3,3) \\\Xcline{2-4}{0.1pt}
       &  \multicolumn{3}{c|}{\small Leaky ReLU}\\\hline
     3 &  16  & 16 & (3,3) \\\Xcline{2-4}{0.1pt}
       &  \multicolumn{3}{c|}{\small Leaky ReLU}\\\hline
     4 &  16  &  2 & (3,3) \\\Xcline{2-4}{0.1pt}
       &  \multicolumn{3}{c|}{$\tanh$}\\\hline\hline
\end{tabular}
    \caption{Convolutional neural network architecture used in coupling layers.}
    \label{tab-cnn}
\end{table}

The python code for the single update step is presented in the listing~\ref{lst-train-step}. Function \lstinline{train_step} is parameterized by the \lstinline{loss_fn} function which  implements the loss used to calculate estimators $\g_i$. The code for loss estimators is presented in listings \ref{lst-delta-1-2} and \ref{lst-delta-3}. The \lstinline{reverse_apply_flow} and \lstinline{apply_flow} functions are presented in listing~\ref{lst-reverse}. 

\begin{lstlisting}[float, caption={A single update step.}, captionpos=b, numbers=none, numberstyle=\tiny, label={lst-train-step}]
def train_step(sub_mean, batch_size, 
               *, model, action, loss_fn, optimizer):
    optimizer.zero_grad()
    loss, logq, logp = loss_fn(sub_mean, model=model, 
                               action=action)
    loss.backward()
    optimizer.step()
\end{lstlisting}

\begin{lstlisting}[float, caption={Loss for estimators $\g_1$ and $\g_2$. They only difference is the subtraction of mean from the signal in case of estimator $\g_2$. \lstinline{layers} implements the affine coupling layers normalizing flow, \lstinline{prior} implements the $q_{pr}(\bz)$ distribution}, captionpos=b, numbers=none, numberstyle=\tiny, label={lst-delta-1-2}]
def g_1_2_loss(sub_mean, batch_size, 
                   *, model, action):
    layers, prior = model["layers"], model["prior"]

    with torch.no_grad():
        with autocast(enabled=use_amp):
            z = prior.sample_n(batch_size)
            log_p_z = prior.log_prob(z) 
            phi, logq = nf.apply_flow(layers, z, log_p_z)
            logp = -action(phi)
            signal =  
        
        z, log_q_phi = nf.reverse_apply_flow(layers, 
                                             phi,
                                             torch.zeros(batch_size,
                                             device=phi.device))
        log_q_phi+= prior.log_prob(z)
        if sub_mean:
            loss = torch.mean(log_q_phi * (signal - signal.mean()))
        else:
            loss = torch.mean(log_q_phi * signal )
            
    return loss, logq, logp
\end{lstlisting}

\begin{lstlisting}[float, caption={Loss for estimator $\g_3$. \lstinline{sub\_mean} parameter is provided for compatibility with $\g_2$ and $\g_3$ loss implementation.}, captionpos=b, numbers=none, numberstyle=\tiny, label={lst-delta-3}]
def g_3_loss(sub_mean, batch_size, 
                 *, model, action):
    layers, prior = model["layers"], model["prior"]
    x, logq = nf.apply_flow_to_prior(
                        prior, layers, 
                        batch_size=batch_size)
    logp = -action(x)
    loss = torch.mean(logq-logp)
    return loss, logq, logp
\end{lstlisting}

\begin{lstlisting}[float, caption={Applying the flow in forward and in reverse directions.}, captionpos=b, label={lst-reverse}]
def apply_flow(coupling_layers, z, logq):
    for layer in coupling_layers:
        z, logJ = layer.forward(z)
        logq = logq - logJ
    return z, logq

def apply_flow_to_prior(prior, coupling_layers, *, batch_size):
    z = prior.sample_n(batch_size)
    logq = prior.log_prob(z)
    return apply_flow(coupling_layers, z, logq)

def reverse_apply_flow(coupling_layers, phi, logq):
    for layer in reversed(coupling_layers):
        phi, logJ = layer.reverse(phi)
        logq += logJ
    return phi, logq
\end{lstlisting}

\begin{table}
    \centering
    \begin{threeparttable}
    \begin{tabular}{|ccc|}\hline\hline
         L = 16 & $m^2=-4$ & $\lambda=8$\\\hline    
         optimizer & Adam & lr = $0.001$\\\hline
        num. epochs & steps per epoch & batch size\\
        4000      & 100           & 1024 \\\hline
           GPU & \multicolumn{1}{c}{V100} &\multicolumn{1}{l|}{32GB} \\\hline
         estimator &   time & t/step\\\hline
         $\g_2$    &  19:00:00  & 0.17s\\
         $\g_3$    &  17:10:00  & 0.15s \\\hline \hline
    \end{tabular}
    \end{threeparttable}
    \caption{\label{tab:runs} Parameters and timings of the runs. We have omitted the $\g_1$ estimator because of its poor performance.}
\end{table}

\subsection{Results}

For each estimator we have made four different training runs of 4000 epochs each, where each epoch consisted of  100  simulation steps and in each step we have sampled  a batch of 1024 $\bphi$ configurations. The training parameters as well as timings are presented in Table~\ref{tab:runs}. 
As expected the $\g_2$ estimator is slower as it has to make one more pass through the network. However it is only $\sim 10\%$ slower, indicating that it is the gradient calculation that takes up most of the time. 

\begin{figure}
    \begin{center}
    \includegraphics[width=0.5\linewidth]{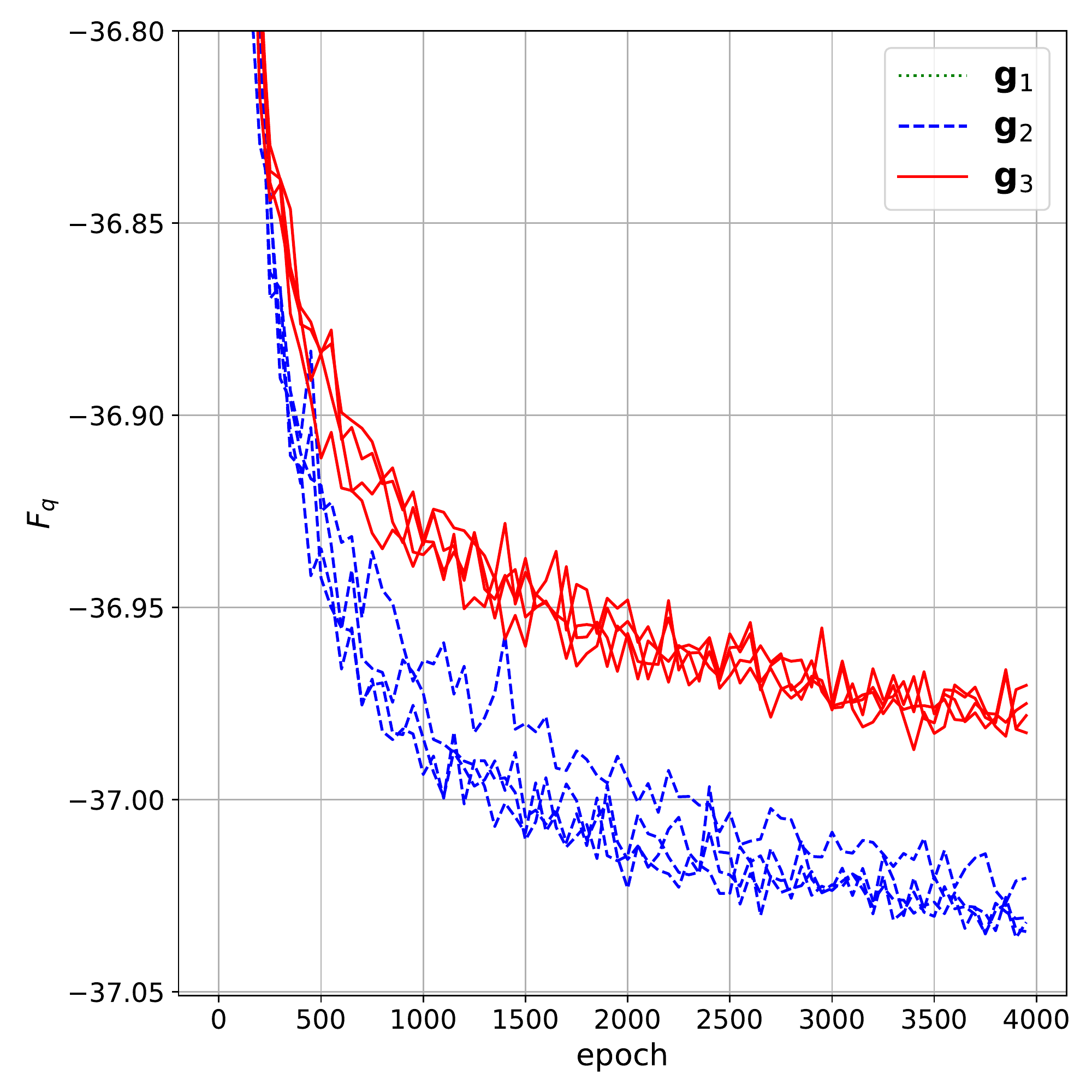}\includegraphics[width=0.5\linewidth]{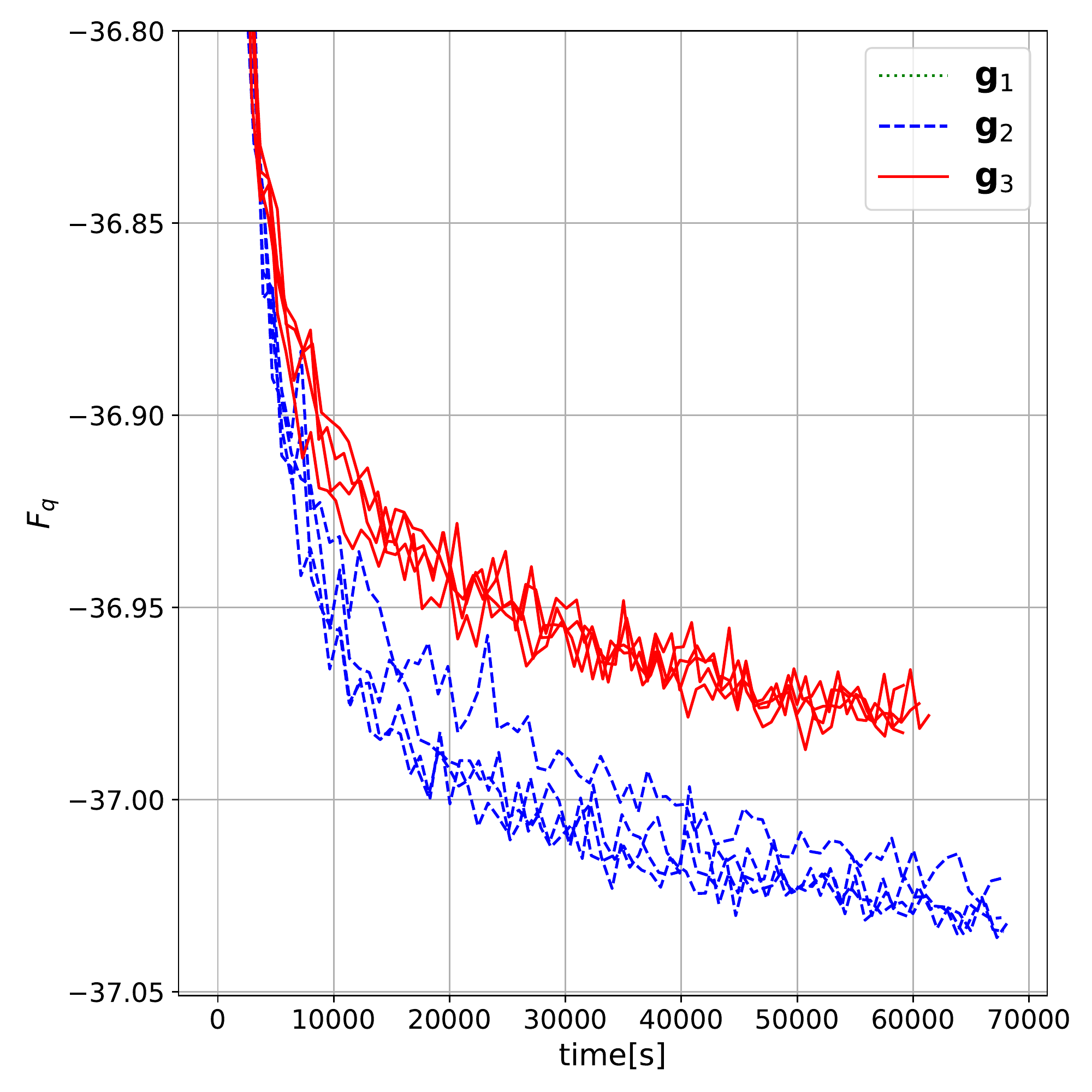}
    \end{center}
    \caption{\label{fig:phi4-dkl}(Left) $F_q$ as the function of epoch. One epoch consisted of 100 steps. In each step the \lstinline{train_step} function was called once.  (Right) $F_q$ as a function of wall time in seconds. The estimator $\g_1$ is not shown as it results in $F_q$ greater then $-36.0$.}
\end{figure}

In Figs.~\ref{fig:phi4-ess} and \ref{fig:phi4-dkl} we present the evolution of two metrics: effective sample size (ESS) and  variational free energy $F_q$ (Eq.~\eqref{eq:shifted-DKL}). 
The ESS is defined as
\begin{equation}
ESS = \frac{\E{w(\bphi)}^2_{q(\bphi|\btheta)}}
{\E{w(\bphi)^2}_{q(\bphi|\btheta)}}
\approx\frac{\left(\sum_{i=1}^N w(\bphi_i)\right)^2}{N\sum_{i=1}^N w(\bphi_i)^2}
\end{equation}    
where 
\begin{equation}
    w(\bphi) =\frac{p(\bphi)}{q(\bphi|\btheta)} \quad\text{and}\quad \bphi_i\sim q(\bphi_i|\btheta)
\end{equation}
and is  an estimate of the fraction of samples that can be considered independent \cite{Liu}. Obviously $\qpt=p(\bphi)$ entails $w(\bphi_i)=1$ and $ESS=1$.
The values presented are the averages over one epoch. First thing one can notice  is that the estimator $\g_1$ is converging very slowly and we have decided to stop training after 1000 epochs and do not consider this estimator in further studies.
Estimators $\g_2$ and $\g_3$ achieve comparable results with estimator $\g_2$ systematically converging faster. 

To better compare the two estimators for each run we  have looked at last 10 epochs (1000 steps) to find the lowest achieved value of $F_q$. We have then saved the  corresponding model.
Each model was then used to generate  a  sample of $10^5$ $\bphi$ configurations. We used those samples to calculate the variational free energy $F_q$ (eq.~\eqref{eq:shifted-DKL}). We have calculated the standard deviation of the signal $s(\bphi|\btheta)-\overline{s(\bphi|\btheta)_N}$ over each sample. This can be used as an indicator of the  quality of training  as it is zero when $q(\bphi|\btheta)=p(\bphi)$. While it is not clear how to quantify this, we can assume that lower  standard deviation indicates better trained flow \cite{VANPRL}. 

Next, we used the Metropolis-Hastings rejection step \eqref{eq:ims} to obtain the Monte-Carlo samples  from distribution \eqref{eq:phi4-prob}. We have calculated the acceptance  and the integrated autocorrelation time $\tau$ (see \cite{Sokal1997} pages 137, 143-145). The results for each run are presented in Table~\ref{tab:est-comp}. For the estimator $\g_2$ we have used models obtained after training for 2000 epochs or 4000 epochs. Looking at the table we see that $\g_2$ systematically outperforms $\g_3$ for every metric even for much shorter training times.

\begin{table}
    \centering
   \begin{tabular}{||rrrr||rrrr||rrrr||}\hline\hline
   \multicolumn{12}{c}{estimator(epochs) time[hh:mm]}\\\hline
\multicolumn{4}{||c||}{$\mathbf{g}_2(2000)$ 9:30} & \multicolumn{4}{c||}{$\mathbf{g}_2(4000)$ 19:00} & \multicolumn{4}{c||}{$\mathbf{g}_3(4000)$ 17:10}\\ \hline
$F_q$ & $\sqrt{var}$ & acc. & $\tau$ & $F_q$ & $\sqrt{var}$ & acc. & $\tau$ & $F_q$ & $\sqrt{var}$ & acc. & $\tau$\\ \hline\hline
-36.98 & 0.76 & 0.69 & 1.28 & -37.03 & 0.70 & 0.73 & 1.09 & -36.96 & 0.77 & 0.66 & 1.43\\ 
-37.00 & 0.74 & 0.70 & 1.21 & -37.02 & 0.71 & 0.72 & 1.09 & -36.97 & 0.76 & 0.67 & 1.41\\ 
-37.01 & 0.72 & 0.71 & 1.18 & -37.04 & 0.68 & 0.74 & 1.03 & -37.00 & 0.74 & 0.69 & 1.38\\ 
-36.98 & 0.76 & 0.67 & 1.29 & -37.04 & 0.68 & 0.75 & 1.00 & -36.99 & 0.74 & 0.69 & 1.27\\ 
\hline
\end{tabular}
    \caption{var denotes the variance of the signal $s(\bphi|\btheta)-\overline{s(\bphi|\btheta)_N}$  and acpt. the acceptance, $\tau$  is the integrated autocorrelation time. }
    \label{tab:est-comp}
\end{table}

All those three estimators are (practically) unbiased,  so the differences in performance must stem from the difference of higher moments. To verify this we estimated the variance of each estimator. Given a model we have generated $N_b=1000$ batches $\{\bphi\}_i$ of 1024 samples each (that was the batch size used in training).  For each batch we have calculated the gradient estimate and calculated the variance of every term which we then averaged 
\begin{equation}\label{eq:grad-var}
   \var{\g} \approx \frac{1}{N_\theta}\sum_{j=1}^{N_\theta}\frac{1}{N_b}\sum_{i=1}^{N_b} 
   \left(g_{j}[\{\bphi\}_i]- \overline{g_j}\right)^2
\end{equation}
where $\g$ is any of three gradients estimators and $g_j[\{\bphi\}]$ is its $j$-th component calculated  for batch $\{\bphi\}$,   
$\overline{g_j}$ is the average of this component over all batches. 

\begin{figure}
    \begin{center}
    \includegraphics[width=0.5\linewidth]{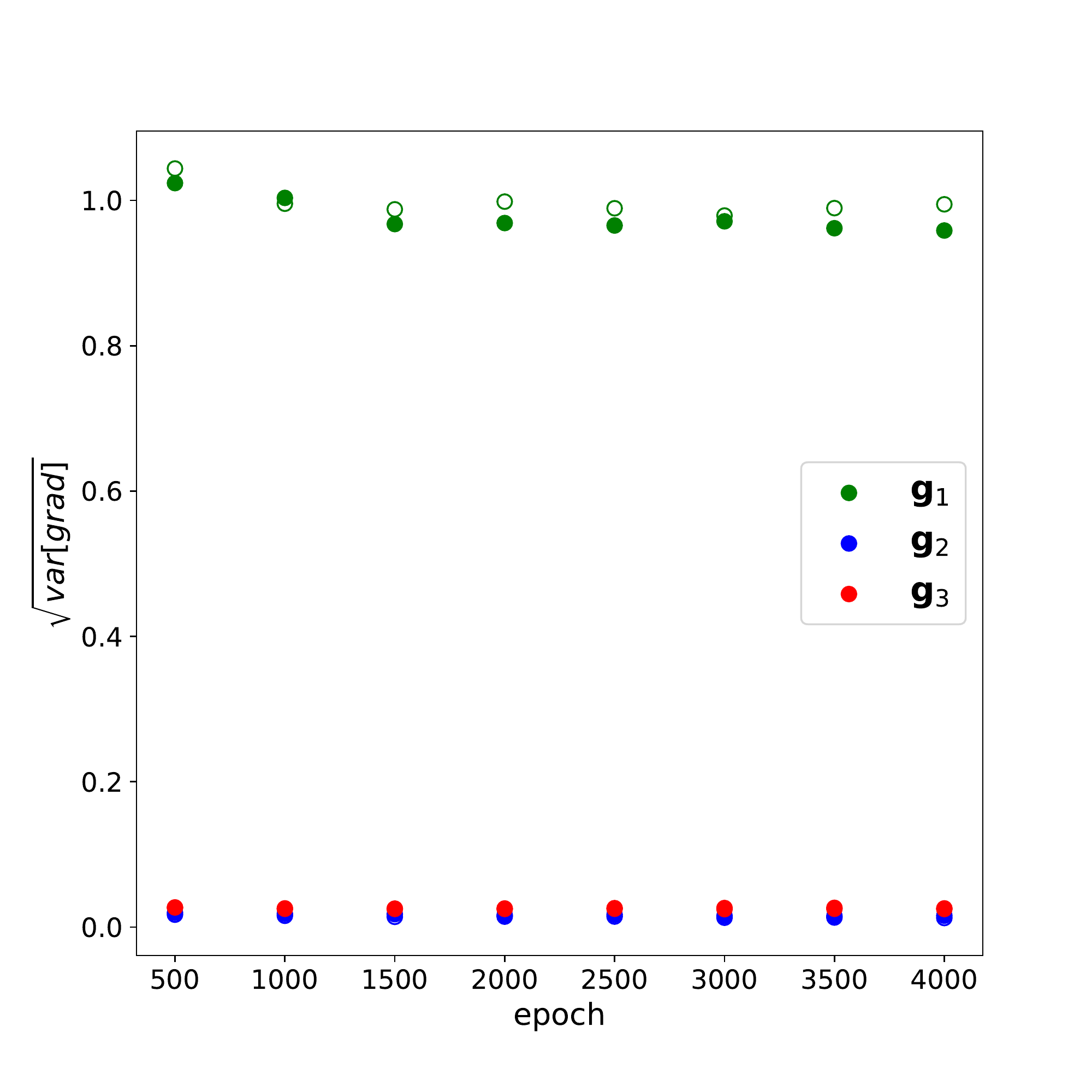}\includegraphics[width=0.5\linewidth]{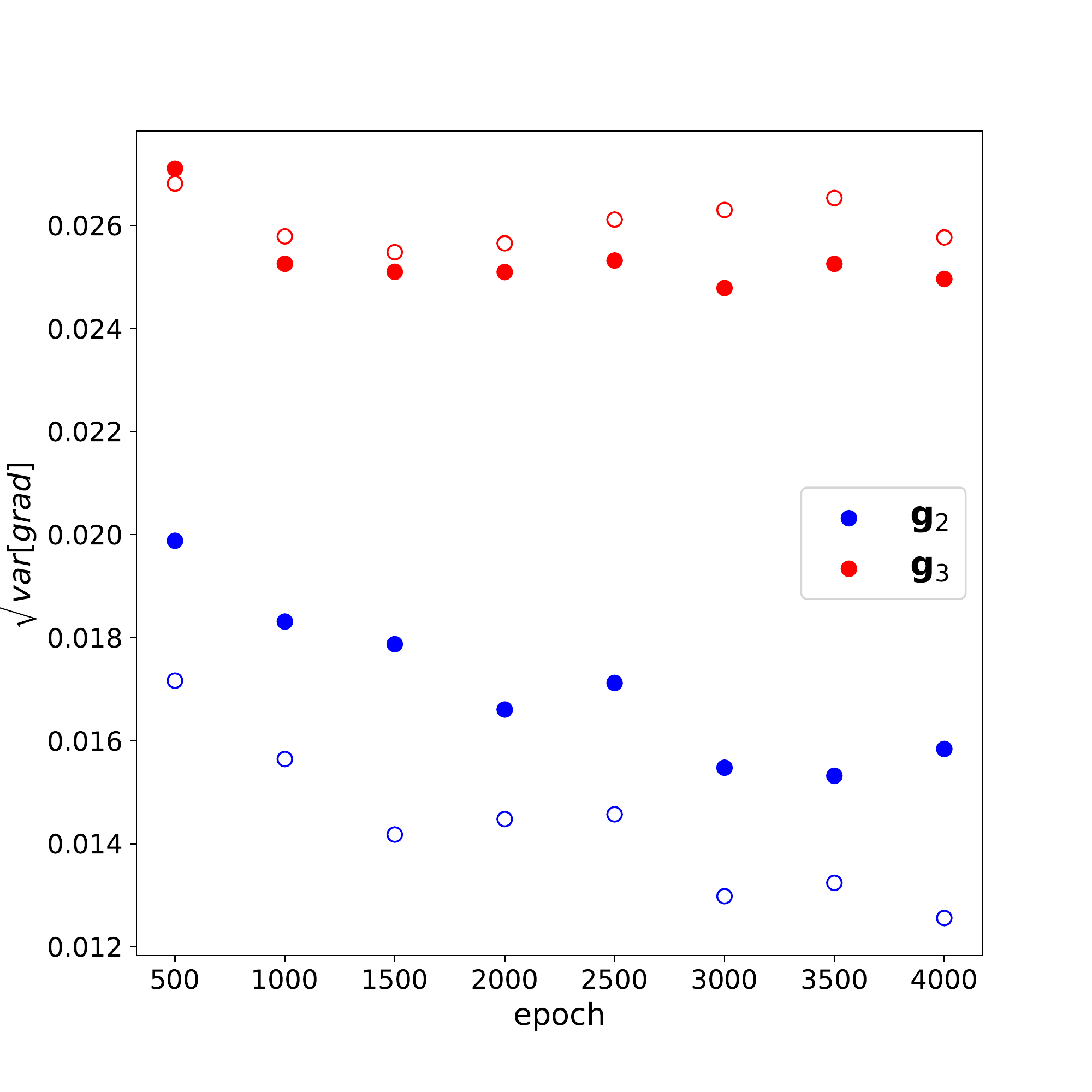}
    \end{center}
    \caption{\label{fig:phi4-grad} Square root of variance of gradient estimators as the function of training time. Right panel shows same data but on different vertical scale. Open and closed symbols show independent runs. }
\end{figure}

The results are presented in the Figure~\ref{fig:phi4-grad}. We plot the square root of variance \eqref{eq:grad-var} as the function of training time of the model. As we can see the values for estimator $\g_1$ are almost two orders of magnitude larger then for other two. That explains why it is performing so poorly. The picture on the right shows same  data but on different vertical scale so we can see the difference between $\g_2$ and $\g_3$ estimators. Estimator $\g_2$ has clearly a lower variance which explains its better performance. 

\section{Summary}

In this work we have described two estimators of the gradient of the loss function used in the Literature in two different contexts: one is usually discussed together with systems with discrete degrees of freedom (we denoted it $\g_2$ in the text), while the other together with systems with continuous degrees of freedom (called $\g_3$). The machine learning architectures used for these two classes of systems are also different: one uses autoregressive neural networks in the first case, while normalizing flows in the second case. We pointed out that the two gradient estimators differ conceptually, namely the estimator typically used in the context of normalizing flows requires an explicit computation of derivatives of the action with respect to the fields, while the other does not. We therefore described how to adapt the gradient estimator $\g_2$ to the case of normalizing flows, rendering the computation of action derivatives unnecessary. This has the potential of speeding up the training for models with more complicated actions. We supplemented our discussion with numerical experiments: in a one-dimensional toy model where all relevant quantities have been calculated analytically as well as in the two-dimensional scalar $\phi^4$ field theory model. The proposed  estimator $\g_2$ takes only $10\%$ more time to calculate, this is more then offset by its better convergence properties. We have shown that given the same training time it can outperform the standard estimator by a large margin and provide similar results in half of the time. Our results suggest that this is due to the lower variance of the $\g_2$ estimator compared to $\g_3$. 

It should be noted that the training of the flow can be regarded as estimating the free energy by a variational approach. Such task is notoriously hard, error prone and time consuming in classical MCMC, this fact can warrant the use of normalising flows  for $F_q$ calculation even when using them in NMCMC may be unpractical \cite{ETOinLFT}. We have shown that using our approach we have obtained lower, and thus better, values of $F_q$. 

\section*{Acknowledgment}
Computer time allocation  grant  \texttt{plgnnformontecarlo} on the Prometheus supercomputer hosted by AGH Cyfronet in Kraków, Poland was used through the polish PLGRID consortium. T.S. kindly acknowledges support of the Polish National Science Center (NCN) Grant No.\,2019/32/C/ST2/00202 and support of the Faculty of Physics, Astronomy and Applied Computer Science, Jagiellonian University Grant No.\,2021-N17/MNS/000062. This research was partially funded by the Priority Research Area Digiworld under the program Excellence Initiative – Research University at the Jagiellonian University in Kraków.

\appendix 

\section{Bias of estimator $\g_2$}
\label{ap:g2}

First, recall that (eq. \ref{eq:signal})
\begin{equation*}
 s(\bphi|\btheta) \equiv  \log q(\bphi|\btheta)-\log P(\bphi)\quad\text{and}\quad  \overline{s(\bphi)_N} =\frac{1}{N}\sum_{i=1}^{N} s(\bphi_i).
\end{equation*}
From \eqref{eq:def-g2} we have
\begin{equation}\label{eq-grad-KL-delta-2}
    \E{\g_2[\{\bphi\}]}= \E{\g_1[\{\bphi\}]}-\frac{1}{N}\sum_{i=1}^N\E{\delta(\bphi|\btheta)\overline{s(\bphi|\btheta)_N}},
\end{equation}
where we have introduced a shortened notation
\begin{equation}
    \delta(\bphi|\btheta) = \diffp{ \log q(\bphi|\btheta)}{\btheta}.
\end{equation}
The second term in the  expression \eqref{eq-grad-KL-delta-2} is equal to 
\begin{equation}
\begin{split}
    \E{\frac{1}{N^2}\sum_{i,j} \delta(\bphi_i|\btheta)s(\bphi_j|\btheta)}&=\frac{1}{N}\E{\delta(\bphi|\btheta)s(\bphi|\btheta)}+\frac{N-1}{N}\E{\delta(\bphi|\btheta)}\E{s(\bphi|\btheta)}=\frac{1}{N}\E{\g_1[\{\bphi\}]}
\end{split}
\end{equation}
where we have used the formula \eqref{eq-const-der} which entails $\E{\delta(\bphi)}=0$. Putting this back into \eqref{eq-grad-KL-delta-2} we obtain the stated result \eqref{eq:g2-bias}. 

\section{Variance of the estimators}
\label{ap:variance}

Because $\bphi$'s are independent from \eqref{eq-KL-grad-approx} we obtain
\begin{equation}
    \var{\g_1}=\frac{1}{N}\left(\E{ \delta(\bphi|\btheta)^2s(\bphi|\btheta)^2}-\E{ \delta(\bphi|\btheta) s(\bphi|\btheta)}^2\right),
 \end{equation}
when $q(\bphi|\btheta)=p(\bphi)$ then $s(\bphi|\btheta)=-\log Z$ and we obtain expression \eqref{eq:g1-variance}.

For estimator $\g_2$ by the same reasoning we obtain
\begin{equation}
     \var{\g_2}=\frac{1}{N}\left(\E{ \delta(\bphi|\btheta)^2 \left(s(\bphi|\btheta)-\overline{s(\bphi|\btheta)_N} \right)^2}-\E{ \delta(\bphi|\btheta)  \left(s(\bphi|\btheta)-\overline{s(\bphi|\btheta)_N} \right)}^2\right),
\end{equation}
when $q(\bphi|\btheta)=p(\bphi)$ then $s(\bphi|\btheta)-\overline{s(\bphi|\btheta)_N} =0$ and variance vanishes.

\bibliography{nmcmc}

\begin{thebibliography}{18}
\expandafter\ifx\csname natexlab\endcsname\relax\def\natexlab#1{#1}\fi
\providecommand{\url}[1]{\texttt{#1}}
\providecommand{\href}[2]{#2}
\providecommand{\path}[1]{#1}
\providecommand{\DOIprefix}{doi:}
\providecommand{\ArXivprefix}{arXiv:}
\providecommand{\URLprefix}{URL: }
\providecommand{\Pubmedprefix}{pmid:}
\providecommand{\doi}[1]{\href{http://dx.doi.org/#1}{\path{#1}}}
\providecommand{\Pubmed}[1]{\href{pmid:#1}{\path{#1}}}
\providecommand{\bibinfo}[2]{#2}
\ifx\xfnm\relax \def\xfnm[#1]{\unskip,\space#1}\fi
\bibitem[{Metropolis et~al.(1953)Metropolis, Rosenbluth, Rosenbluth, Teller,
  and Teller}]{metropolis}
\bibinfo{author}{N.~Metropolis}, \bibinfo{author}{A.~W. Rosenbluth},
  \bibinfo{author}{M.~N. Rosenbluth}, \bibinfo{author}{A.~H. Teller},
  \bibinfo{author}{E.~Teller},
\newblock \bibinfo{title}{Equation of state calculations by fast computing
  machines},
\newblock \bibinfo{journal}{The Journal of Chemical Physics}
  \bibinfo{volume}{21} (\bibinfo{year}{1953}) \bibinfo{pages}{1087--1092}.
  \URLprefix \url{https://doi.org/10.1063/1.1699114}.
  \DOIprefix\doi{10.1063/1.1699114}.
  \href{http://arxiv.org/abs/https://doi.org/10.1063/1.1699114}{{\tt
  arXiv:https://doi.org/10.1063/1.1699114}}.
\bibitem[{Binder and Heermann(2019)}]{binder}
\bibinfo{author}{K.~Binder}, \bibinfo{author}{D.~Heermann},
  \bibinfo{title}{{Monte Carlo Simulation in Statistical Physics: An
  Introduction}}, \bibinfo{publisher}{Springer}, \bibinfo{year}{2019}.
\bibitem[{Wu et~al.(2019)Wu, Wang, and Zhang}]{VANPRL}
\bibinfo{author}{D.~Wu}, \bibinfo{author}{L.~Wang}, \bibinfo{author}{P.~Zhang},
\newblock \bibinfo{title}{Solving statistical mechanics using variational
  autoregressive networks},
\newblock \bibinfo{journal}{Phys. Rev. Lett.} \bibinfo{volume}{122}
  (\bibinfo{year}{2019}) \bibinfo{pages}{080602}.
\bibitem[{Nicoli et~al.(2020)Nicoli, Nakajima, Strodthoff, Samek, M\"uller, and
  Kessel}]{PhysRevE.101.023304}
\bibinfo{author}{K.~A. Nicoli}, \bibinfo{author}{S.~Nakajima},
  \bibinfo{author}{N.~Strodthoff}, \bibinfo{author}{W.~Samek},
  \bibinfo{author}{K.-R. M\"uller}, \bibinfo{author}{P.~Kessel},
\newblock \bibinfo{title}{Asymptotically unbiased estimation of physical
  observables with neural samplers},
\newblock \bibinfo{journal}{Phys. Rev. E} \bibinfo{volume}{101}
  (\bibinfo{year}{2020}) \bibinfo{pages}{023304}.
\bibitem[{Białas et~al.(2021)Białas, Korcyl, and Stebel}]{bialas2021analysis}
\bibinfo{author}{P.~Białas}, \bibinfo{author}{P.~Korcyl},
  \bibinfo{author}{T.~Stebel}, \bibinfo{title}{Analysis of autocorrelation
  times in neural markov chain monte carlo simulations}, \bibinfo{year}{2021}.
  \href{http://arxiv.org/abs/2111.10189}{{\tt arXiv:2111.10189}}.
\bibitem[{Paszke et~al.(2019)}]{PyTorch}
\bibinfo{author}{A.~Paszke}, et~al.,
\newblock \bibinfo{title}{Pytorch: An imperative style, high-performance deep
  learning library},
\newblock in: \bibinfo{booktitle}{Advances in Neural Information Processing
  Systems 32}, \bibinfo{publisher}{Curran Associates, Inc.},
  \bibinfo{year}{2019}, pp. \bibinfo{pages}{8024--8035}.
\bibitem[{Liu(1996)}]{Liu}
\bibinfo{author}{J.~S. Liu},
\newblock \bibinfo{title}{Metropolized independent sampling with comparisons to
  rejection sampling and importance sampling},
\newblock \bibinfo{journal}{Statistics and Computing} \bibinfo{volume}{6}
  (\bibinfo{year}{1996}) \bibinfo{pages}{113--119}.
\bibitem[{Nicoli et~al.(2021)}]{ETOinLFT}
\bibinfo{author}{K.~A. Nicoli}, et~al.,
\newblock \bibinfo{title}{Estimation of thermodynamic observables in lattice
  field theories with deep generative models},
\newblock \bibinfo{journal}{Physical Review Letters} \bibinfo{volume}{126}
  (\bibinfo{year}{2021}). \URLprefix
  \url{http://dx.doi.org/10.1103/PhysRevLett.126.032001}.
  \DOIprefix\doi{10.1103/physrevlett.126.032001}.
\bibitem[{Dinh et~al.(2017)Dinh, Sohl-Dickstein, and Bengio}]{dinh2017density}
\bibinfo{author}{L.~Dinh}, \bibinfo{author}{J.~Sohl-Dickstein},
  \bibinfo{author}{S.~Bengio}, \bibinfo{title}{Density estimation using real
  nvp}, \bibinfo{year}{2017}. \href{http://arxiv.org/abs/1605.08803}{{\tt
  arXiv:1605.08803}}.
\bibitem[{Albergo et~al.(2019)Albergo, Kanwar, and
  Shanahan}]{PhysRevD.100.034515}
\bibinfo{author}{M.~S. Albergo}, \bibinfo{author}{G.~Kanwar},
  \bibinfo{author}{P.~E. Shanahan},
\newblock \bibinfo{title}{Flow-based generative models for markov chain monte
  carlo in lattice field theory},
\newblock \bibinfo{journal}{Phys. Rev. D} \bibinfo{volume}{100}
  (\bibinfo{year}{2019}) \bibinfo{pages}{034515}.
\bibitem[{Kobyzev et~al.(2020)Kobyzev, Prince, and Brubaker}]{9089305}
\bibinfo{author}{I.~Kobyzev}, \bibinfo{author}{S.~Prince},
  \bibinfo{author}{M.~Brubaker},
\newblock \bibinfo{title}{Normalizing flows: An introduction and review of
  current methods},
\newblock \bibinfo{journal}{IEEE Transactions on Pattern Analysis and Machine
  Intelligence}  (\bibinfo{year}{2020}) \bibinfo{pages}{1--1}.
\bibitem[{Frey(1998)}]{graphical_models}
\bibinfo{author}{B.~J. Frey}, \bibinfo{title}{Graphical Models for Machine
  Learning and Digital Communication}, \bibinfo{publisher}{MIT Press,
  Cambridge, MA}, \bibinfo{year}{1998}.
\bibitem[{Uria et~al.(2016)Uria, C{{\^o}}t{{\'e}}, Gregor, Murray, and
  Larochelle}]{NeuralAutoregressive}
\bibinfo{author}{B.~Uria}, \bibinfo{author}{M.-A. C{{\^o}}t{{\'e}}},
  \bibinfo{author}{K.~Gregor}, \bibinfo{author}{I.~Murray},
  \bibinfo{author}{H.~Larochelle},
\newblock \bibinfo{title}{Neural autoregressive distribution estimation},
\newblock \bibinfo{journal}{Journal of Machine Learning Research}
  \bibinfo{volume}{17} (\bibinfo{year}{2016}) \bibinfo{pages}{1--37}.
\bibitem[{Germain et~al.(2015)Germain, Gregor, Murray, and
  Larochelle}]{MaskedAutoencoders}
\bibinfo{author}{M.~Germain}, \bibinfo{author}{K.~Gregor},
  \bibinfo{author}{I.~Murray}, \bibinfo{author}{H.~Larochelle},
\newblock \bibinfo{title}{Made: Masked autoencoder for distribution
  estimation},
\newblock in: \bibinfo{editor}{F.~Bach}, \bibinfo{editor}{D.~Blei} (Eds.),
  \bibinfo{booktitle}{Proceedings of the 32nd International Conference on
  Machine Learning}, volume~\bibinfo{volume}{37} of
  \textit{\bibinfo{series}{Proceedings of Machine Learning Research}},
  \bibinfo{publisher}{PMLR}, \bibinfo{address}{Lille, France},
  \bibinfo{year}{2015}, pp. \bibinfo{pages}{881--889}.
\bibitem[{Oord et~al.(2016)Oord, Kalchbrenner, and Kavukcuoglu}]{pixelCNN}
\bibinfo{author}{A.~V. Oord}, \bibinfo{author}{N.~Kalchbrenner},
  \bibinfo{author}{K.~Kavukcuoglu},
\newblock \bibinfo{title}{Pixel recurrent neural networks},
\newblock in: \bibinfo{editor}{M.~F. Balcan}, \bibinfo{editor}{K.~Q.
  Weinberger} (Eds.), \bibinfo{booktitle}{Proceedings of The 33rd International
  Conference on Machine Learning}, volume~\bibinfo{volume}{48} of
  \textit{\bibinfo{series}{Proceedings of Machine Learning Research}},
  \bibinfo{publisher}{PMLR}, \bibinfo{address}{New York, New York, USA},
  \bibinfo{year}{2016}, pp. \bibinfo{pages}{1747--1756}.
\bibitem[{Abadi et~al.(2015)}]{tensorflow2015-whitepaper}
\bibinfo{author}{M.~Abadi}, et~al., \bibinfo{title}{{TensorFlow}: Large-scale
  machine learning on heterogeneous systems}, \bibinfo{year}{2015}. \URLprefix
  \url{https://www.tensorflow.org/about/bib}, \bibinfo{note}{software available
  from \texttt{tensorflow.org}}.
\bibitem[{Albergo et~al.(2021)}]{albergo2021introduction}
\bibinfo{author}{M.~S. Albergo}, et~al., \bibinfo{title}{Introduction to
  normalizing flows for lattice field theory}, \bibinfo{year}{2021}.
  \href{http://arxiv.org/abs/2101.08176}{{\tt arXiv:2101.08176}}.
\bibitem[{Sokal(1997)}]{Sokal1997}
\bibinfo{author}{A.~Sokal},
\newblock \bibinfo{title}{Monte carlo methods in statistical mechanics:
  Foundations and new algorithms},
\newblock in: \bibinfo{editor}{C.~DeWitt-Morette},
  \bibinfo{editor}{P.~Cartier}, \bibinfo{editor}{A.~Folacci} (Eds.),
  \bibinfo{booktitle}{Functional Integration: Basics and Applications},
  \bibinfo{publisher}{Springer US}, \bibinfo{address}{Boston, MA},
  \bibinfo{year}{1997}, pp. \bibinfo{pages}{131--192}.

\end{thebibliography}

\end{document}